\documentclass[twoside]{article}

\usepackage[accepted]{aistats2020}

\usepackage[round]{natbib}

\title{Sampling Acquisition Functions for Batch Bayesian Optimization}

\usepackage[noend]{algpseudocode}
\usepackage{algorithm, algorithmicx}

\usepackage{amsmath, amssymb}
\DeclareMathOperator*{\argmin}{arg\,min}
\DeclareMathOperator*{\argmax}{arg\,max}
\usepackage{hyperref}
\usepackage[noend]{algpseudocode}

\usepackage{graphicx}
\usepackage{booktabs}
\usepackage{enumitem}
\usepackage{cleveref}
\usepackage{dblfloatfix}  
\usepackage{siunitx}  
\usepackage[font=footnotesize,labelfont=bf]{caption}

\begin{document}

\runningauthor{A. De Palma, C. Mendler-D{\"u}nner, T. Parnell, A. Anghel, H. Pozidis}

\twocolumn[

\aistatstitle{Sampling Acquisition Functions for Batch Bayesian Optimization}

\aistatsauthor{ Alessandro De Palma \And {Celestine Mendler-D{\"u}nner} }
\aistatsaddress{University of Oxford \And UC Berkeley  } 
\aistatsauthor{Thomas Parnell, Andreea Anghel, Haralampos Pozidis }
\aistatsaddress{ IBM Research, Zurich } 
 ]

\begin{abstract}
  We present Acquisition Thompson Sampling (ATS), a novel technique for batch Bayesian Optimization (BO) based on the idea of sampling multiple acquisition functions from a stochastic process. We define this process through the dependency of the acquisition functions on a set of model hyper-parameters. ATS is conceptually simple, straightforward to implement and, unlike other batch BO methods, it can be employed to parallelize any sequential acquisition function or to make existing parallel methods scale further.
  We present experiments on a variety of benchmark functions and on the hyper-parameter optimization of a popular gradient boosting tree algorithm.
  These demonstrate the advantages of ATS with respect to classical parallel Thompson Sampling for BO, its competitiveness with two state-of-the-art batch BO methods, and its effectiveness if applied to existing parallel BO algorithms.
\end{abstract}

\section{Introduction}

Bayesian optimization (BO) methods deal with the optimization of expensive, black-box functions, i.e., functions with no closed-form expression or derivative information. Many such problems arise in practice in fields such as: hyper-parameter optimization of machine learning algorithms \citep{spearmint}, synthesis of short polymer fiber materials \citep{Li2017}, and analog circuit design \citep{lyu18a}. Therefore, in recent years, BO gained widespread popularity to the point that its versatility was showcased on toy problems like optimizing chocolate cookie recipes \citep{cookies}.

\newpage

From a high-level perspective, BO works by training a probabilistic regression model (typically, Gaussian Processes) on the available function evaluations. This model is then employed to decide where to evaluate the function next, maximizing a trade-off between exploration and exploitation that is formalized through an acquisition function $a(\boldsymbol{x})$. Many sequential BO algorithms exist, differing by the choice of the regression model, the acquisition function, or properties of the input domain \citep{SMAC, Srinivas12, shahriari16, Springerberg16, hernandez-lobatob15}.
Recently, parallel BO techniques have gained popularity. These methods select multiple points to be evaluated in parallel in order to better exploit modern hardware capabilities, and to speed up the overall optimization process in the case of costly evaluations like training machine learning algorithms. In the parallel case, typically, the acquisition function must return $q$ points at once (batch BO), these being greedily selected one after the other.

Most successful batch BO algorithms focus on parallelizing a specific acquisition function \citep{kandasamy18a, pkg, Daxberger17, ppes, gupta18a}. In this paper, instead, we will focus on methods that can be applied to any sequential acquisition function, as in general no single one consistently outperforms the others \citep{Hoffman2011}.
Amongst these general methods, the prevalent approach is to modify the landscape of the acquisition function $a(\boldsymbol{x})$ after a point is scheduled so as to avoid returning another point in its neighborhood. For instance, Local Penalization (LP) \citep{Gonzalez} accomplishes this by suppressing (penalizing) the value of $a(\boldsymbol{x})$ in the neighborhood of any scheduled $\boldsymbol{x}$ using an approximation of the Lipschitz constant of the unknown function. A family of algorithms including, e.g., B-LCB \citep{GP-BUCB}, instead, obtains the same effect by \textit{hallucinating} the results of the points that have been scheduled but not yet completed.

\subsection{Intuition, Contributions, Challenges}
In general, a sequential acquisition function depends on a vector of parameters mostly belonging to the employed regression model. We will denote the parameter vector by $\boldsymbol{\theta}$, and write $a(\boldsymbol{x}) = a(\boldsymbol{x}, \boldsymbol{\theta})$ to refer to a member of the family of acquisition functions. The $\boldsymbol{\theta}$ vector exerts a strong effect on the practical performance of BO \citep{spearmint}. Hence, in practice, it is either set to its maximum likelihood value, or its influence is mitigated by using an acquisition function averaged over a fixed number of possible values of $\boldsymbol{\theta}$.
Our intuition is that, rather than focusing on modifying the landscape of a fixed $a(\boldsymbol{x}, \{\boldsymbol{\theta}_i\})$, where $\{\boldsymbol{\theta}_i\}$ possibly denotes a set of vectors, such a dependency can be exploited to select each point in a batch from a different member of the family. In order to do so, inspired by Parallel Thompson Sampling (P-TS) \citep{kandasamy18a}, we interpret a given sequential acquisition function $a(\boldsymbol{x}, \{\boldsymbol{\theta}_i\})$ as being sampled from a stochastic process.
It then suffices to draw samples from the process to obtain as many different acquisition functions as the number of batch points to evaluate.
Our contributions are the following: \vspace{-10pt}
\begin{itemize}[leftmargin =8pt, itemsep=0.05\baselineskip]
	\item We present ATS, a novel batch Bayesian optimization algorithm based on the idea of sampling acquisition functions from a stochastic process defined by their dependency on hyper-parameters $\boldsymbol{\theta}$. ATS is conceptually simple and easy to implement. In contrast to the previously published randomized approach \citep{kandasamy18a}, ATS can parallelize any arbitrary sequential acquisition function and can be employed to make existing parallel acquisition functions scale further. 
	\item We show how existing techniques to diversify points within a batch, can be incorporated into our randomized framework to improve performance of ATS on highly multi-modal functions.
	\item We present results on five well-known benchmark functions and on hyper-parameter tuning for XGBoost \citep{KDD16_xgboost}. When applied to sequential acquisition functions, we demonstrate the advantages of ATS compared to P-TS~\citep{kandasamy18a}, B-LCB~\citep{GP-BUCB} and LP~\citep{Gonzalez}, three popular state-of-the-art batch BO methods. In particular, ATS consistently finds better minima than P-TS.
	\item We show that ATS can be successfully used to improve the performance of existing parallel methods, P-TS and B-LCB, on our benchmark functions.
\end{itemize}
\vspace{-10pt}
We leave the formulation of regret bounds as future work. Most regret bounds (sequential and parallel), such as those for LCB ~\citep{Srinivas12}, B-LCB~\citep{GP-BUCB}, TS~\citep{chowdhury17a}, P-TS~\citep{kandasamy18a} allow only fixed hyper-parameters \citep{loop}. In practice, hyper-parameters are changed dynamically in the optimization (set to a point estimate) or approximately marginalized out. The former approach might not converge to the global minimum \citep{Bull} unless, as it was very recently shown, further constraints are imposed on the hyper-parameters \citep{Berkenkamp19}. We employ marginalization: to the best of our knowledge, a regret bound is missing even under a fully Bayesian treatment for sequential BO \citep{loop}. In addition, we would need to prove that hyper-parameter diversity speeds up parallel convergence. Therefore, we believe that regret bounds for ATS would be an interesting and challenging direction for future work also in the purely sequential context.

\section{Problem Statement and Background}

Our goal is to find the minimum of a black-box function $f: \mathcal{X} \rightarrow \mathbb{R}$, defined on a bounded set $\mathcal{X} \subset \mathbb{R}^d$. While the function itself is unknown, we have access to the results of its evaluations $f(\boldsymbol{x})$ at any domain point $\boldsymbol{x}$. We are interested in scenarios where these evaluations are \emph{costly} and we seek to minimize $f$ using as few evaluations as possible. For instance, $f$ might represent the loss incurred by a machine learning algorithm trained with the hyper-parameter vector $\boldsymbol{x}$.

\subsection{Bayesian Optimization} \label{sec:related-sequential}

Bayesian Optimization (BO) approaches this black-box optimization problem by placing a prior on $f$ and updating the model by conditioning on the evaluations as they are collected.
Once the \emph{surrogate} model is updated, BO employs it to select the next point to evaluate according to a criterion referred to as the \emph{acquisition function}. This function should not only \textit{exploit} the regions that are likely to contain the minimum, but also \textit{explore} those where uncertainty is significant.

Gaussian Processes (GP) are a common and versatile choice for the function prior. Under a GP, any finite number of function outputs is jointly distributed according to a multivariate Gaussian distribution $\mathcal{N}(\boldsymbol{\mu}, K)$,
where the covariance $K$ of the Gaussian process is defined through a kernel function $k$, parametrized by $\boldsymbol{\theta}_k$ as $K_{i,j}=k(\boldsymbol{x}_i, \boldsymbol{x}_j | \boldsymbol{\theta}_k)$. The choice of the kernel function $k$ determines the family of functions that a GP can represent. We use the Mat\'ern 5/2 kernel due to its widespread usage and its good empirical performance on tuning machine learning algorithms \citep{spearmint}. Moreover, we assume the prior mean $\boldsymbol{\mu}_p$ to be constant (yet possibly non-zero) on the input domain. In general, while the function $f$ is deterministic, its evaluations may be noisy. It is, therefore, common practice to assume Gaussian noise in addition to the GP, so that $y_i = f(\boldsymbol{x}_i) + \mathcal{N}(0, \sigma_{n}^2 I)$.
We will denote the set of collected observations up to evaluation $t$ with $\mathcal{D}_t = \{(\boldsymbol{x}_i, y_i))\ |\ i < t\}$.
A desirable property of GPs in this context is that $\mathcal{N}(\boldsymbol{\mu}, K | \mathcal{D}_t)$ is again a GP, and its posterior distribution at a point can be computed in closed form in $\mathcal{O}(t^3)$ time \citep{GPBook}.
In the remainder of the paper, we will ignore the evaluation noise and set $y_i = f(\boldsymbol{x}_i)$ to facilitate the presentation. In order to generalize to the noisy case, it suffices to substitute $f(\boldsymbol{x}_i)$ with $y_i$ and to include the noise in the kernel matrix.
Once $t$ evaluations are available, the next point is selected by maximizing the acquisition function: $\boldsymbol{x}_{t+1} = \argmax_{\mathcal{X}} a(\boldsymbol{x}|\mathcal{D}_t)$, where the conditioning explicitly denotes the dependency on collected evaluations $\mathcal{D}_t$ through the updated surrogate model. Two widespread sequential acquisition functions are Expected Improvement (EI) \citep{JonesSW98} and Lower Confidence Bound (LCB) \citep{Srinivas12}:
\begin{eqnarray*}
\text{EI}(\boldsymbol{x}|\mathcal{D}_t) &=& E_{f(\boldsymbol{x} | \mathcal{D}_t)}[\max (f(\boldsymbol{x}^-) - f(\boldsymbol{x}), 0) ] \\
\text{LCB}(\boldsymbol{x}|\mathcal{D}_t) &=& \sigma(\boldsymbol{x}|\mathcal{D}_t) - \mu(\boldsymbol{x}|\mathcal{D}_t)
\end{eqnarray*}
where $f(\boldsymbol{x}^-)$ denotes the best (smallest) evaluation result recorded until evaluation $t$.

\subsection{Parallel Bayesian Optimization} \label{sec:related-parallel}

Bayesian Optimization as presented in the previous section is an inherently sequential algorithm: points are evaluated one by one, and the model is updated to reflect the evaluation results. Many works have explored the possibility of evaluating more than one point at once, either asynchronously \citep{spearmint, kandasamy18a} or synchronously \citep{Gonzalez, GP-BUCB, gupta18a, DPP, lyu18a, Daxberger17}. Next, we will focus on the latter, more studied, case, as asynchronous algorithms can also be employed in a synchronous fashion.
Some approaches, like $q$-EI by \citet{qEI}, theoretically generalize sequential acquisition functions to the case in which $q$ points must be returned at once. Unfortunately, such approaches are not computationally feasible for $q>4$, as the batch of points needs to be jointly optimized \citep{Daxberger17}. Therefore, most of the attention has been devoted to greedy methods, where points are selected one after the other. ATS falls within this category.
We will call \emph{pending evaluations}, the points for which an evaluation has been scheduled but not yet completed. We will denote all pending evaluations with the prime sign: if evaluation $i$ is pending, we will write $\boldsymbol{x}_i'$, meaning that $\boldsymbol{x}_i'$ is available, but not $f(\boldsymbol{x}_i')$.

A plethora of approaches for greedy batch BO exists \citep{Gonzalez, GP-BUCB, gupta18a, DPP, lyu18a, Daxberger17, spearmint, kandasamy18a}; here, we will describe those that form the background to our algorithms.
A well-known heuristic is to \emph{hallucinate} the results of the pending evaluations, i.e., to approximate them with known quantities. This way, the acquisition function is updated from $a(\boldsymbol{x}|\mathcal{D}_t)$ to $a(\boldsymbol{x}|\mathcal{D}_t \cup \{(\boldsymbol{x}_i', h_i)\ \forall\ \boldsymbol{x}_i' \})$, where $h_i$ denotes the hallucinated evaluation result for the pending evaluation $\boldsymbol{x}_i'$. For instance, the very effective batch LCB (B-LCB) approach updates the surrogate model after each batch point has been scheduled by setting $f(\boldsymbol{x}_i') =\mu(\boldsymbol{x}_*|\mathcal{D}_t)$ \citep{GP-BUCB}.
The batch BO algorithm by \citet{gupta18a}, instead, parametrizes the LCB acquisition function to express different explore-exploit trade-offs and, arguing that different trade-offs are associated to different maxima of $a(\boldsymbol{x}|\mathcal{D}_t)$, assigns each batch point a new trade-off.

\subsection{Thompson Sampling} \label{sec:ts}

Thompson Sampling (TS) \citep{thompson1933} is a rather old heuristic to address the explore-exploit dilemma that gained popularity in the context of multi-armed bandits \citep{Thompson}. The general idea is to choose an action according to the probability that it is optimal.
Applied to BO, this corresponds to sampling a function $g$ from the GP posterior and selecting $\boldsymbol{x}_{t+1} = \argmax_{\mathcal{X}} g$ \citep{kandasamy18a}.
In fact, let $p_{*}(\boldsymbol{x})$ be the probability that point (action) $\boldsymbol{x}$ optimizes the GP posterior, then:
\begin{align}
p_{*}(\boldsymbol{x})  &= \int p_{*}(\boldsymbol{x} | g)\ p(g | \mathcal{D}_t)\ dg \notag\\
&= \int \delta(\boldsymbol{x} - \argmin_{\mathcal{X}} g(\boldsymbol{x}))p(g | \mathcal{D}_t)\ dg.
 \label{eq:ts}
\end{align}
where $\delta$ denotes the Dirac delta distribution.
TS is therefore equivalent to a randomized sequential acquisition function, one that is easy to parallelize. In fact, it suffices to sample multiple functions from the GP posterior to get a batch of points to be evaluated in parallel (P-TS) \citep{kandasamy18a}.
In this paper, we generalize the reasoning in \eqref{eq:ts} so as to parallelize an arbitrary acquisition function, and additionally incorporate ideas from hallucinations and intra-batch trade-off differentiation.

\section{Sampling Acquisition Functions}

We first introduce \emph{Acquisition Thompson Sampling} (ATS), our novel algorithm for both batch and asynchronous parallel Bayesian Optimization (\S\ref{sec:thompson-mcmc}) and then discuss two extensions thereof, j-ATS and h-ATS (\S\ref{sec:hallucination}). The common idea behind these methods is to sample an acquisition function from a stochastic process for each parallel evaluation to perform.

\subsection{Acquisition Thompson Sampling (ATS)} \label{sec:thompson-mcmc}
Any acquisition function $a(\boldsymbol{x})$ explicitly depends on the surrogate model's hyper-parameters $\boldsymbol{\theta}$ and can be therefore written as $a(\boldsymbol{x},\boldsymbol{\theta})$.

\paragraph{Sequential treatment of $\boldsymbol{\theta}$.}
The most common way to choose $\boldsymbol{\theta}$ is to optimize the marginal likelihood under the Gaussian process. As acquisition functions are very sensitive to these hyper-parameters, \citet{spearmint} suggest the adoption of a fully Bayesian approach by marginalizing over them:
\begin{equation}
a(\boldsymbol{x} | \mathcal{D}_t) = \int a(\boldsymbol{x}, \boldsymbol{\theta}| \mathcal{D}_t)\ p(\boldsymbol{\theta} | \mathcal{D}_t)\ d\boldsymbol{\theta}
\label{eq:int}
\end{equation}
As the exact marginalization in \eqref{eq:int} is intractable, it is approximated by $\tilde{a}_s(\boldsymbol{x})$, the empirical average of $s$ samples from the data posterior of $\boldsymbol{\theta}$ obtained via Markov Chain Monte Carlo (MCMC) sampling \citep{spearmint}:
\begin{equation}\label{eq:as}
\tilde{a}_s(\boldsymbol{x} | \mathcal{D}_t) = \frac{1}{s} \sum_{q=1}^{s} a(\boldsymbol{x}, \boldsymbol{\theta}_q | \mathcal{D}_t) \text{ s.t. } \boldsymbol{\theta}_q \sim p(\boldsymbol{\theta} | \mathcal{D}_t)
\end{equation}
\paragraph{Exploiting $\boldsymbol{\theta}$ for parallelism.}
In this paper, we aim to leverage the sensitivity to $\boldsymbol{\theta}$ for selecting a diverse batch of points to be evaluated in parallel. In order to do so, we propose an alternative interpretation of \eqref{eq:as}. The main intuition is to recognize that, for low $s$, $\tilde{a}_s(\boldsymbol{x})$ represents a sample from a probability distribution over functions that is implicitly defined by the $\boldsymbol{\theta}$ data posterior, i.e., a stochastic process. We will denote such a process by $a_s(\boldsymbol{x} | \mathcal{D}_t)$, and write $\tilde{a}_s \sim p(\tilde{a}_s | \mathcal{D}_t)$ to refer to the sampling in \eqref{eq:as}. We can then argue that the sequential criterion for determining the next point to evaluate,
$\boldsymbol{x}_{t+1} = \argmax_{\mathcal{X}} \tilde{a}_s(\boldsymbol{x} | \mathcal{D}_t)$,
can be seen as sampling from $p_{a_s^*}(\boldsymbol{x})$, the probability that $\boldsymbol{x}$ maximizes the stochastic process. We can write:
\begin{align}
\label{eq:ats}
p_{a_s^*}(\boldsymbol{x}) &= \int p_{a_s^*}(\boldsymbol{x} | \tilde{a}_s)\ p(\tilde{a}_s | \mathcal{D}_t)\ d\tilde{a}_s \\
&= \int \delta(\boldsymbol{x} - \argmax_{\mathcal{X}} \tilde{a}_s(\boldsymbol{x}| \mathcal{D}_t)) p(\tilde{a}_s | \mathcal{D}_t)\ d \tilde{a}_s \notag
\end{align}
\begin{algorithm}[t]
	\caption{Batch iteration, Acquisition Thompson Sampling (ATS).}\label{alg:thompson-mcmc}
	\begin{algorithmic}[1]
		\Require dataset $\mathcal{D}_{t}$, acquisition function to parallelize $a(\boldsymbol{x}, \boldsymbol{\theta}| \mathcal{D}_t)$, batch size $M>0$, $s>0$.
		\For{$i = 1, \dots, M$ \textbf{in parallel}}
		\State \parbox[t]{\dimexpr\linewidth-\algorithmicindent} {sample $s$ new GP hyper-parameter vectors according to $\boldsymbol{\theta}_q \sim p(\boldsymbol{\theta} | \mathcal{D}_t)$. \strut}
		\State \parbox[t]{\dimexpr\linewidth-\algorithmicindent} {$\tilde{a}_s(\boldsymbol{x} | \mathcal{D}_t) \leftarrow \frac{1}{s} \sum_{q=1}^{s} a(\boldsymbol{x}, \boldsymbol{\theta}_q | \mathcal{D}_t)$. \strut}
		\State \parbox[t]{\dimexpr\linewidth-\algorithmicindent} {$\boldsymbol{x}_{t+i} \leftarrow \argmax_{\mathcal{X}} \tilde{a}_s(\boldsymbol{x} | \mathcal{D}_t)$\strut} 
		\State \parbox[t]{\dimexpr\linewidth-\algorithmicindent} {evaluate $f(\boldsymbol{x}_{t+i})$. \strut}
		\EndFor

		\State $\mathcal{D}_{t + M} \leftarrow \mathcal{D}_{t} \cup \{(\boldsymbol{x}_k, f(\boldsymbol{x}_k))\ \forall\ t < k \leq t+M\}\ $.
	\end{algorithmic}
\end{algorithm}
In light of this interpretation, a sequential acquisition function can be employed in the parallel setting by sampling as many different acquisition functions from $a_s(\boldsymbol{x} | \mathcal{D}_t)$ (i.e., as many approximations of the marginalized function $a(\boldsymbol{x} | \mathcal{D}_t)$) as the parallel evaluations to perform. We call this procedure ATS as
our algorithm closely resembles Thompson Sampling for Bayesian Optimization (cf. \S \ref{sec:ts}). In fact, \eqref{eq:ats} can be interpreted as an adaptation of \eqref{eq:ts} to a case in which the optimal action is defined with respect to a stochastic acquisition function rather than the GP posterior. While P-TS is an acquisition function of its own, this adaptation allows us to obtain a general approach, applicable for the parallelization of \emph{any} sequential acquisition function by means of its parametrization $\boldsymbol{\theta}$.
In Section \ref{sec:experiments}, we demonstrate empirically that the $\tilde{a}_s$ samples are diverse enough to provide significant speed-ups to sequential BO, and that ATS consistently finds better minima than P-TS.
Algorithm \ref{alg:thompson-mcmc} outlines the pseudo-code for a single iteration of the batch version of ATS. For implementation details, including the chosen prior for $\boldsymbol{\theta}$ and the employed MCMC sampling algorithm, cf. Appendix \ref{app:details}.
\paragraph{The role of $s$.}
The parameter $s$ plays a fundamental role in ATS, as it controls the trade-off between sample diversity and surrogate model reliability. In fact, $s \rightarrow \infty$ yields perfect marginalization and no parallelism, while $s=1$ implies significant batch diversity but a poor representation of the black-box function $f$.
Therefore, $s$ should be chosen to be a monotonically decreasing function of the batch size $M$.
In practice, we found $s=10$ to be a good compromise between sample diversity and surrogate model reliability for our target functions and levels of parallelism (cf. $\S$~\ref{sec:experiments}).
\paragraph{Enhancing parallel acquisition functions.}
Due to its generality, ATS can be also employed on top of an existing greedy parallel approach to make it scale further. In fact, $a(\boldsymbol{x}, \boldsymbol{\theta}| \mathcal{D}_t)$ in Algorithm \ref{alg:thompson-mcmc} could represent batch BO acquisition functions like P-TS or B-LCB. For the latter, in addition to hallucinating the surrogate model, one would sample new hallucinated functions from the stochastic process for every batch point, with the goal of increasing batch diversity. In order to match the properties of the chosen parallel acquisition function with a variable degree of accuracy, we toss a biased coin for each point, sampling a new function from the process with probability $p$.

\subsection{Techniques for Batch Diversity } \label{sec:j-h-ats}
If $f$ exhibits multiple minima, ATS might over-exploit the current posterior of the function, producing a batch of points centered around a local minimum. We propose two variants of ATS to address this issue.
\paragraph{Explore-exploit trade-offs.}
Pursuing different ex-plore-exploit trade-offs within a batch encourages the exploration of regions of the input domain with varying degrees of posterior uncertainty. This idea was implemented in \citet{gupta18a} by explicitly solving a multi-objective optimization problem on $\text{LCB}_j(\boldsymbol{x} | \mathcal{D}_t) = j \sigma(\boldsymbol{x}|\mathcal{D}_t) - \mu(\boldsymbol{x}|\mathcal{D}_t)$, a parametrized version of LCB, and taking $M$ different trade-offs from the resulting Pareto curve (one per batch point). In this context, the parameter $j$ is commonly called jitter: the larger the jitter, the more explorative the acquisition function.
We build on the intuition of \citet{gupta18a}, but rather than explicitly solving the multi-objective optimization, we translate this idea into our randomized framework by \emph{sampling} different trade-offs from a distribution.
In order to do this, we associate a new process $a_{s,j}(\boldsymbol{x} | \mathcal{D}_t)$ to the parametrization, and place an independent prior on $j$ so as to sample jitter alongside the $\boldsymbol{\theta}_i$ vectors.
In addition to LCB, such a parametrization is commonly employed for many popular acquisition functions; e.g., EI \citep{gpyopt2016}:
\[\text{EI}_j(\boldsymbol{x} | \mathcal{D}_t) = E_{f(\boldsymbol{x} | \mathcal{D}_t)}[\max (f(\boldsymbol{x}^-) - j - f(\boldsymbol{x}), 0) ]\]
As a consequence, sampling from $a_{s,j}(\boldsymbol{x} | \mathcal{D}_t)$ implies obtaining acquisition functions having different explore-exploit trade-offs.
Pseudo-code for a batch iteration of this \emph{jittered} variant ($j$-ATS) of the algorithm, along with details of the employed data posterior for $j$, can be found in Appendix \ref{app:details}.
\paragraph{Hallucinating Pending Evaluations.}\label{sec:hallucination}
The main drawback of $j$-ATS lies in the loss of generality that arises from the need of using an acquisition function-specific prior. As an alternative, we propose to add \emph{hallucinations} of pending evaluations  (see $\S$ \ref{sec:related-parallel}) to the data posterior of $\boldsymbol{\theta}$. This is in contrast with previous work, where hallucinations are added to the surrogate model \citep{GP-BUCB}. The resulting method, $h$-ATS, is fully described in Appendix \ref{app:hallucination}.

\vspace{-7pt}
\section{Experimental Results} \label{sec:experiments}
\vspace{-7pt}
This section presents the results of our empirical evaluations for the proposed ATS algorithm (and its variants), applied both to sequential and parallel acquisition functions. In Section \ref{sec:synthetic-res} we benchmark on well-known synthetic functions from the global optimization literature and in Section \ref{sec:xgb-res} we present the results on the hyper-parameter tuning of a gradient boosting tree algorithm on a binary classification task.

\begin{figure*}[!t]
	\vspace{-10pt}
	\centering
	\hbox{\hspace{-40pt} \includegraphics[width =1.2 \textwidth]{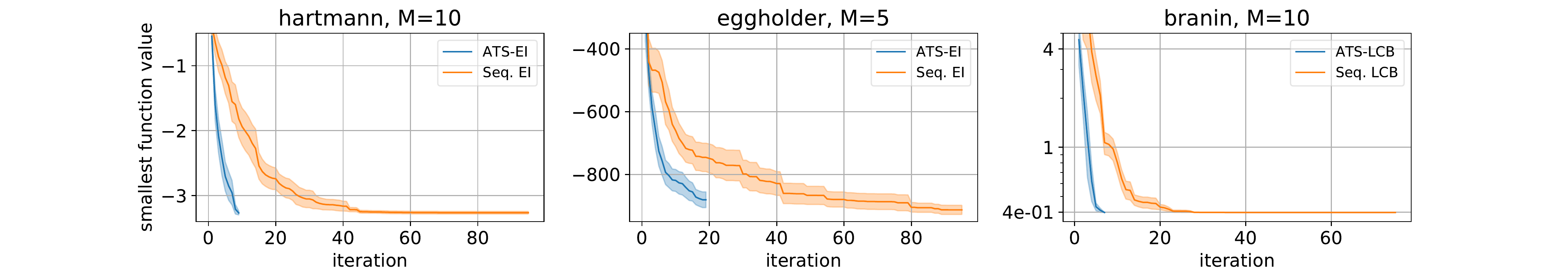} }
	\caption{Parallel vs. sequential BO on synthetic functions for varying batch size $M$ and acquisition function. \vspace{-0.1cm}}
	\label{fig:selection-sequential}
\end{figure*}

\vspace{-8pt}
\paragraph{Methodology \& Baselines.}
We compare with two state-of-the-art, well-known batch Bayesian optimization algorithms: Local Penalization (LP) \citep{Gonzalez} and batch LCB (B-LCB) \citep{GP-BUCB}, and against Parallel "classical" Thompson Sampling (P-TS) \citep{kandasamy18a}. LP was taken from the implementation available within the open-source GPyOpt package \citep{gpyopt2016}. We instead re-implemented B-LCB and P-TS within our framework.
While LP is a general batch BO approach (applicable to any acquisition function), B-LCB was presented and analyzed only for LCB \footnote{Their hallucination technique might generalize, but we adhere to the common experimental practice.} and P-TS parallelizes the TS acquisition function only.
Both B-LCB and LP are very competitive against many of the more complex, non-general algorithms presented in the last two years \citep{Daxberger17, gupta18a, lyu18a, DPP}. Moreover, their performance is superior to other algorithms commonly employed as baselines, such as q-KG \citep{pkg} and q-EI \citep{Chevalier2013}; we hence forgo such a comparison.

We employ two widely used acquisition functions, LCB and EI (cf. Section \ref{sec:related-sequential}) for LP, and implement ATS for EI, LP, B-LCB and P-TS. For ATS applied on top of parallel acquisition functions, we sample a new $a(\boldsymbol{x})$ for each point with probability $p=1$ or $p=0.5$.
For all algorithms, we employ a GP surrogate model with Mat\'ern 5/2 kernel \citep{spearmint} and (except for LP, whose original implementation does not support it) we approximately marginalize the acquisition function hyper-parameters $\boldsymbol{\theta}$ after each batch iteration, as suggested by \citep{spearmint}. We stress that this contrasts with common experimental practice \citep{lyu18a}, which sets $\boldsymbol{\theta}$ to its maximum likelihood value, and with the original algorithm presentations \citep{GP-BUCB,kandasamy18a,Gonzalez}. In practice, marginalization yields a decrease in regret of up to three orders of magnitude (for the Cosines function), leading to much stronger baselines.
For ATS, we re-sample multiple $\boldsymbol{\theta}$s from their data posterior for each batch point, as explained in Section \ref{sec:thompson-mcmc}. For both approximate marginalization and ATS, we perform the $\boldsymbol{\theta}$ sampling by employing the Affine Invariant MCMC Ensemble sampler provided by the \emph{emcee} Python library \citep{emcee}.
We decided to perform all the experiments in the batch BO setting, rather than asynchronous, to keep the uncertainty constant and fixed across the algorithms and time. We remark that in practice ATS and the baselines can be easily adapted to work in the asynchronous setting. Surrogate models are always initialized with $5$ random evaluations (not shown).

\subsection{Synthetic Functions} \label{sec:synthetic-res}
Here, we benchmark on a set of synthetic functions that are notoriously hard to optimize due to the presence of multiple local or global minima (Eggholder, Cosines, Hartmann), or deep valley-like regions (Branin, Rosenbrock). Therefore, they are well-known in the global optimization community, and have been previously employed as benchmarks in the context of batch BO \citep{DPP, Gonzalez, gupta18a, kandasamy18a}. More information on the functions can be found in Table \ref{table-benchmark-functions} in the Appendix.


\vspace{-5pt}
\paragraph{Parallelizing a Sequential Acquisition Function.} Before comparing to the parallel BO baselines, we demonstrate that the diversity induced by sampling surrogate model hyper-parameters alone yields an efficient parallel BO algorithm.
Figure \ref{fig:selection-sequential} compares pure ATS to sequential BO, using LCB or EI, on three qualitatively different synthetic functions, plotting the smallest function value found until batch iteration $t$. Mean values and one standard error over at least $10$ repetitions are reported. ATS consistently outperforms the sequential baseline and achieves close to ideal speedup on the high-dimensional and multi-modal Hartmann function. The speedups on Eggholder and Branin ($3.1\times$ for $M=5$ and $5\times$ for $M=10$, respectively) further highlight the benefits of parallel BO and illustrate the effectiveness of ATS in parallelizing an arbitrary acquisition function.

\vspace{-5pt}
\paragraph{ATS Variants vs. Parallel Baselines.}

\begin{figure*}[!t]
	\vspace{-13pt}
	\centering
	\hbox{\hspace{-10pt} \includegraphics[width =1.1 \textwidth]{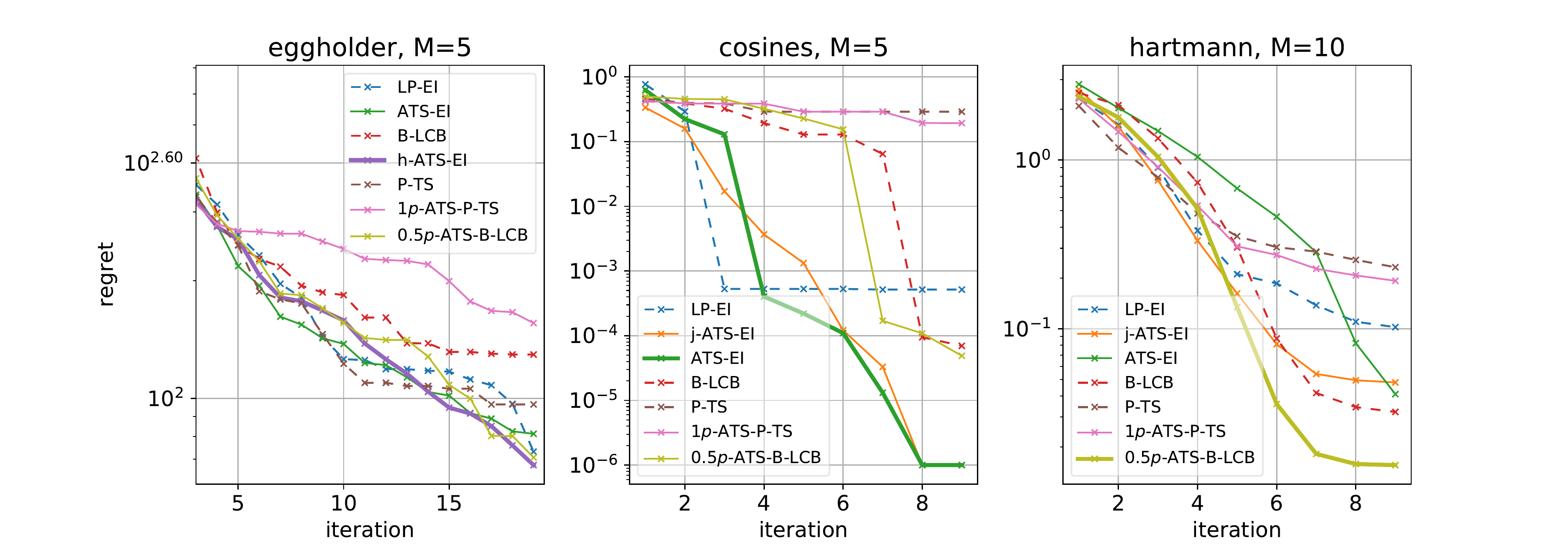} }
	\vspace{-5pt}
	\caption{Benchmark on multi-modal synthetic functions for varying batch size $M$.\vspace{-0.3cm}}
	\label{fig:synthetic-multimodal}
\end{figure*}

Figures \ref{fig:synthetic-valley} and \ref{fig:synthetic-multimodal} present our benchmark results, providing an experimental comparison of the ATS variants to the parallel baselines: LP, B-LCB and P-TS. For LP and ATS, we chose the acquisition function that showed the best sequential behavior\footnote{For Hartmann6, where EI and LCB are on par, we selected EI, as it exhibited a better parallel behavior.} on the given synthetic function. It is in fact well-known that the relative performance of acquisition functions depends on the optimization problem at hand \citep{Hoffman2011}. The plots report mean values of regret (distance from the function minimum, clipped at $\num{1e-6}$) over at least $10$ repetitions of the experiments; the ATS variant with the best performance on the given task is highlighted with a thicker line.
In all plots but one, plain ATS eventually improves on P-TS, highlighting the benefits of focusing on the probability of maximizing an acquisition function rather than the model posterior.

We separate the benchmark functions into two categories:
\begin{figure}[!b]
	\vspace{-20pt}
	\centering
	\hbox{\includegraphics[width =0.55 \textwidth]{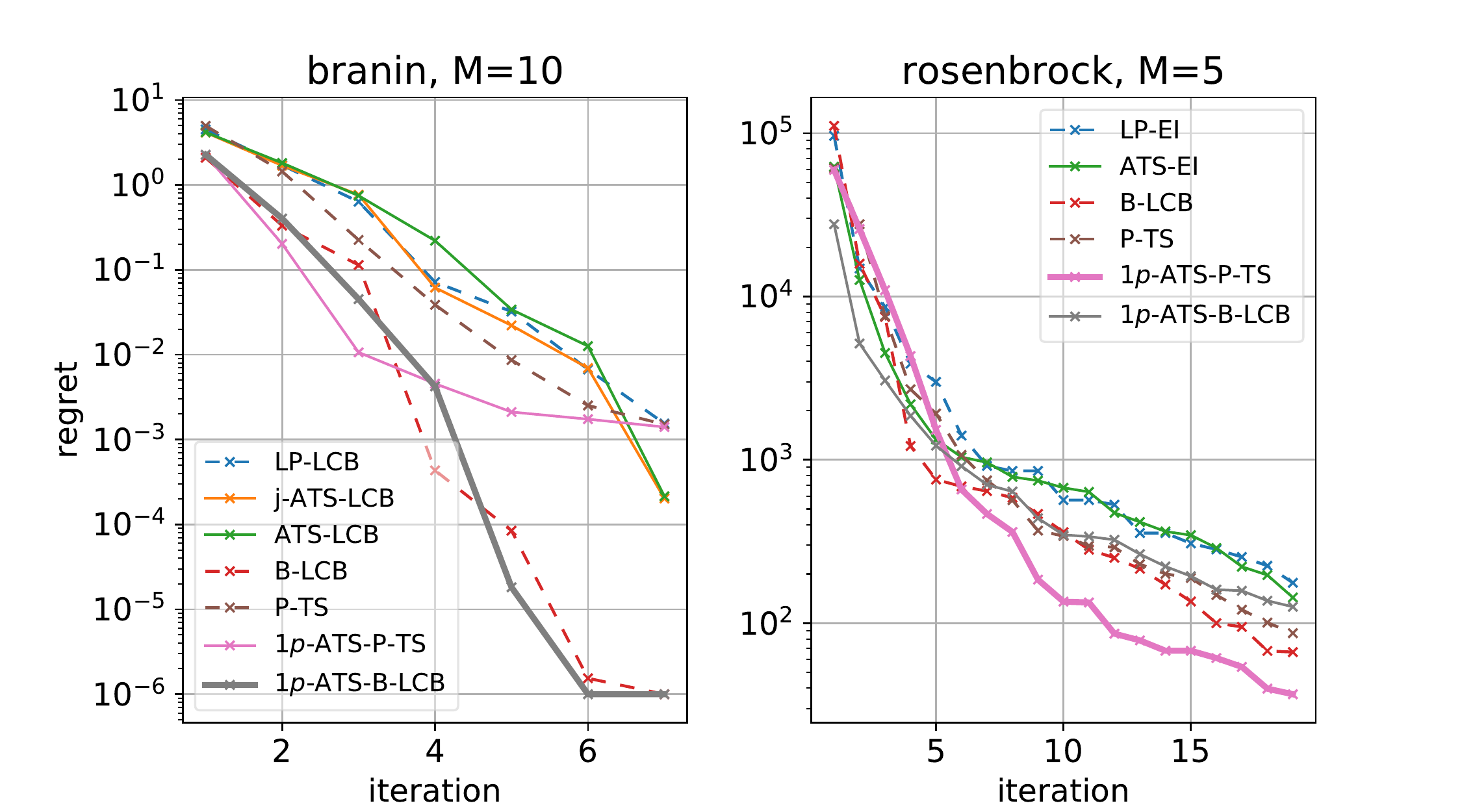} }
	\captionof{figure}{Benchmark on valley-like functions for varying batch size $M$ and acquisition function.}
	\label{fig:synthetic-valley}
\end{figure}
\begin{table*}[t]
\small
	\centering
	\caption{Minima found by batch BO methods after a fixed number of iterations. The table reports mean values over at least $10$ repetitions and one standard error. Column title is structured as: synthetic function, number of batch iterations, batch size, acquisition function. $^\ast$B-LCB and P-TS employ respectively LCB and TS as acquisition functions in all cases.}
	\resizebox{\textwidth}{!}{%
		\begin{tabular}{rccccc}
			\toprule
			Algorithm & Branin & Cosines & Hartmann& Eggholder& Rosenbrock\\
			&7 it., $M=10$, LCB &  9 it., $M=5$, EI & 9 it., $M=10$, EI &  19 it., $M=5$, EI &19 it., $M=5$, EI \\
			\midrule
			Sequential & $1.0682 \pm 0.1666$ & $-1.5187 \pm  0.1117$ & $-1.8188 \pm  0.2862$ & $-745.4952 \pm  45.2170$ & $1630.9149 \pm  352.1795$ \\
			ATS & $0.3981 \pm  0.0002$ & $\mathbf{-1.77321 \pm  \num[math-rm=\mathbf]{6e-8}}$ & $-3.2810 \pm  0.0172$ & $-878.4349 \pm  17.5301$ &  $143.4917 \pm  30.7108$ \\
			LP & $0.3994 \pm  0.0012$ & $-1.7727 \pm \num{1e-5}$& $-3.2196 \pm  0.0293$ & $-886.5274 \pm  14.9021$ &  $176.5940 \pm   42.9050$ \\
			B-LCB$^\ast$ & $0.3979 \pm \num{2e-6}$ & $-1.7731 \pm  \num{5e-5}$ & $-3.2898 \pm 0.0144$ & $-830.1668 \pm  27.8264$ &  $66.4483 \pm  12.6149$ \\
			P-TS$^\ast$ & $0.3994 \pm 0.0003$ & $-1.4832 \pm 0.1476$ & $-3.0904 \pm 0.0261$ & $-863.1183 \pm 39.2462$ &  $86.7537 \pm  28.9031$ \\
		    ATS-B-LCB$^\ast$ & $\mathbf{0.3979 \pm  \num[math-rm=\mathbf]{1e-6}}$ & $-1.77315 \pm  \num{4e-5}$ & $\mathbf{-3.3064 \pm 0.0109}$ & $\mathbf{-888.9844 \pm  17.3145}$ &  $125.8718 \pm 37.7826 $ \\
			ATS-P-TS$^\ast$ & $0.3993 \pm  0.0003$ & $-1.5798 \pm 0.1289$ & $-3.1295 \pm  0.0202$ & $-803.8516 \pm 50.5051$ &  $\mathbf{36.6755 \pm 6.3740}$ \\
			\bottomrule
		\end{tabular}
	}
	\label{table-summary}
\end{table*}
%
Figure \ref{fig:synthetic-valley} presents results for valley-like functions. ATS on sequential methods performs comparably to P-TS and LP while B-LCB seems to excel on low-dimensional, relatively easy to optimize functions like Branin. The use of ATS to enhance existing parallel acquisition functions is successful in this setting: P-TS and B-LCB are almost always improved upon, and enhanced versions are the best performing methods.
On the strongly multi-modal functions shown in Figure \ref{fig:synthetic-multimodal} (Cosines and Eggholder exhibit sinusoidal oscillations), instead, the ATS variants applied to the sequential functions usually already outperform the baselines, implying that the dependence on hyper-parameters alone offers a competitive degree of parallelism.
An exception is the Hartmann function, for which B-LCB performs particularly well, and its ATS enhancement outperforms all other methods.
For these objective functions, diversity seems to be crucial to avoid stagnating in one of the many minima. In fact, the variants of ATS presented in $\S$\ref{sec:j-h-ats} are the algorithms showing the best average performance.
Based on our experiments, of all the analyzed variants, we would recommend the use of ATS-B-LCB or j-ATS. $p=0.5$ is a good trade-off for using ATS on parallel acquisition functions in most use cases.
A summary of the optimization results at the last iteration of the plots in Figures \ref{fig:synthetic-valley} and \ref{fig:synthetic-multimodal}, including the standard error of the mean, can be found in Table \ref{table-summary}.

\begin{figure*}[!t]
	\vspace{-10pt}
	\centering
	\hbox{\hspace{-20pt} \includegraphics[width = 1.1 \textwidth]{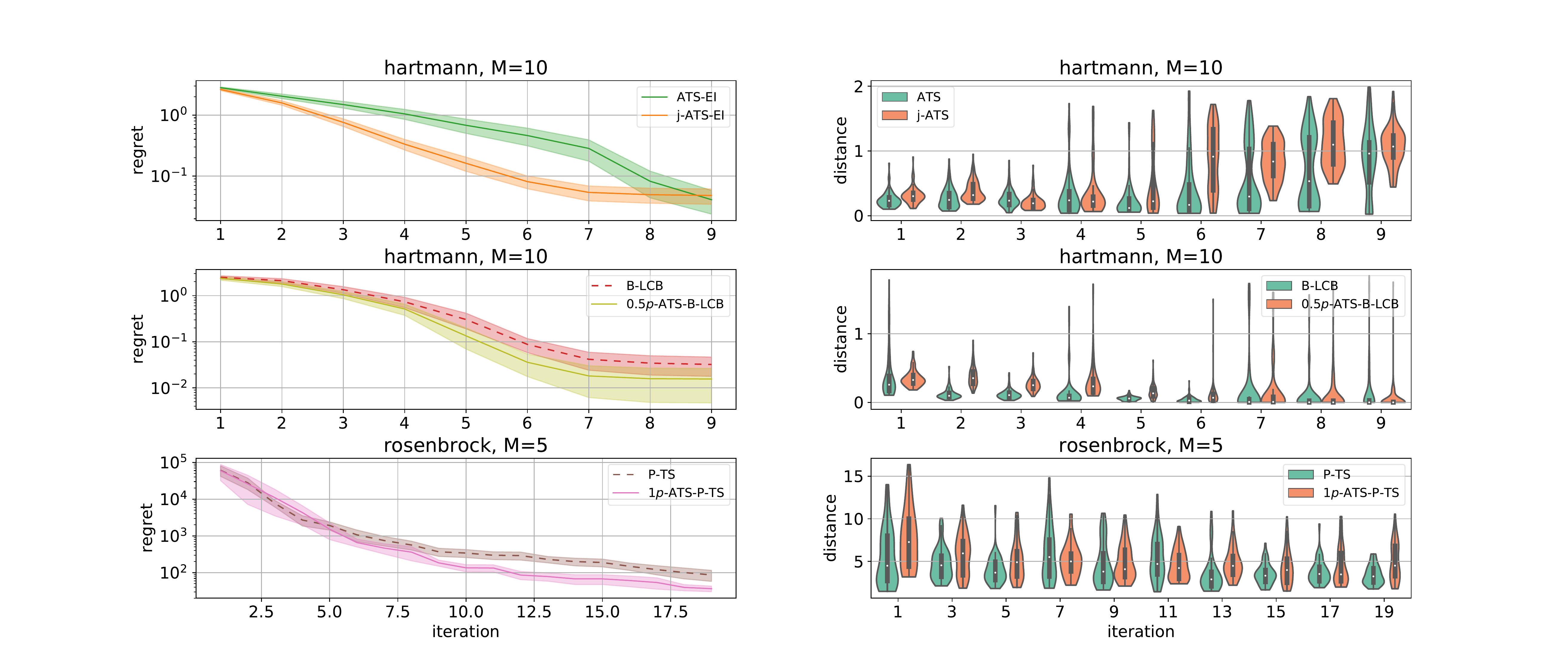} }
	\vspace{-10pt} \caption{Comparison of techniques for batch diversity: regret and intra-batch average $\ell_2$ distance.\vspace{-0.2cm}}
	\label{fig:diversity-plot}
\end{figure*}

\paragraph{Intra-batch analysis.}
In Figure \ref{fig:diversity-plot}, we turn to a more detailed analysis of the ATS variants by looking within the batches. As in Figure \ref{fig:selection-sequential}, we plot the mean and its standard error for at least $10$ repetitions; we show violin plots (where the white dot represents the mean) for the distribution of the average $\ell_2$ distance of each batch points from the others, across the repetitions. The improvement of $j$-ATS on ATS and of the enhanced versions over the parallel baselines is matched by an increase in batch diversity, visible from the higher mean in the violin plots, and by the lack of the lower tail in the distance distribution. The latter effect is particularly important as it means that clusters of nearby points (with distance close to $0$) are removed when using $j$-ATS on sequential acquisition functions and ATS on the parallel ones. The improvement tends to be more visible in the first few and the last few batch iterations.

\vspace{-5pt}
\subsection{Hyper-Parameter Optimization} \label{sec:xgb-res}
\vspace{-5pt}

We perform a similar comparison on the real-world problem of Hyper-Parameter Optimization (HPO). We focus on gradient boosting decision trees, as they enjoy widespread adoption in academia, industry and competitive data science due to their state-of-the-art performance in a wide variety of machine learning tasks \citep{NIPS17_Ke}. Specifically, our goal is to tune the hyper-parameters of the popular XGBoost \citep{KDD16_xgboost}, so as to minimize the logistic loss incurred on a binary classification problem.
We utilized the well-known Higgs \citep{higgs} dataset, freely available within the UCI ML repository, whose task is to predict whether a given process will produce Higgs bosons or not. As the overall number of signals in the original dataset is larger than ten millions, we uniformly subsampled $\approx1'120'000$ examples for the training, and $\approx400'000$ for the validation set to make the experiments feasible on our infrastructure. We tuned $5$ different hyper-parameters; details can be found in Appendix \ref{app:experiments}.
Figure \ref{fig:xgb-20-iters} shows the minimum loss incurred on the fixed validation set after $t$ batch iterations, averaged over $5$ repetitions. We opted for a large batch size ($M=20$) to mimic a realistic scenario where the significant time cost of HPO needs to be amortized through parallelism. Moreover, we chose LCB over EI due to its better sequential performance, as in the previous section. The results show that ATS variants applied to sequential acquisition functions are competitive to B-LCB also on HPO, while LP shows inferior performance. The ATS variant exhibiting the best behavior is, again, $j$-ATS, which achieves almost ideal speedup to sequential BO. This is to be expected, as in larger batches the diversity provided by sampling $\boldsymbol{\theta}$ alone might not be sufficient, and employing hallucinations might be short-sighted with respect to the future model updates.

\section{Conclusions and Future Work}

We presented Acquisition Thompson Sampling (ATS), a novel technique for parallel Bayesian Optimization (BO) based on sampling acquisition functions from a stochastic process through their dependency on the surrogate model hyper-parameters. ATS can be successfully used to parallelize sequential acquisition functions or to make parallel BO methods scale further.
We enhanced this conceptually simple algorithm by translating ideas from the batch BO literature into our randomized framework:
First ($j$-ATS), by adding jitter to the acquisition functions, we sample different explore-exploit trade-offs from a prior distribution embedded in the definition of the process, as opposed to obtaining these trade-offs by solving a multi-objective optimization problem \citep{gupta18a}.
Second ($h$-ATS), we employ hallucinations of the pending evaluations to diversify the data posterior of the surrogate model hyper-parameters, rather than to update the surrogate model itself \citep{GP-BUCB}. We obtain a versatile and multi-faceted algorithm, usable to parallelize any sequential acquisition function.

\begin{figure}[!b]
\vspace{-0.5cm}
	\centering
	\includegraphics[width=0.45 \textwidth]{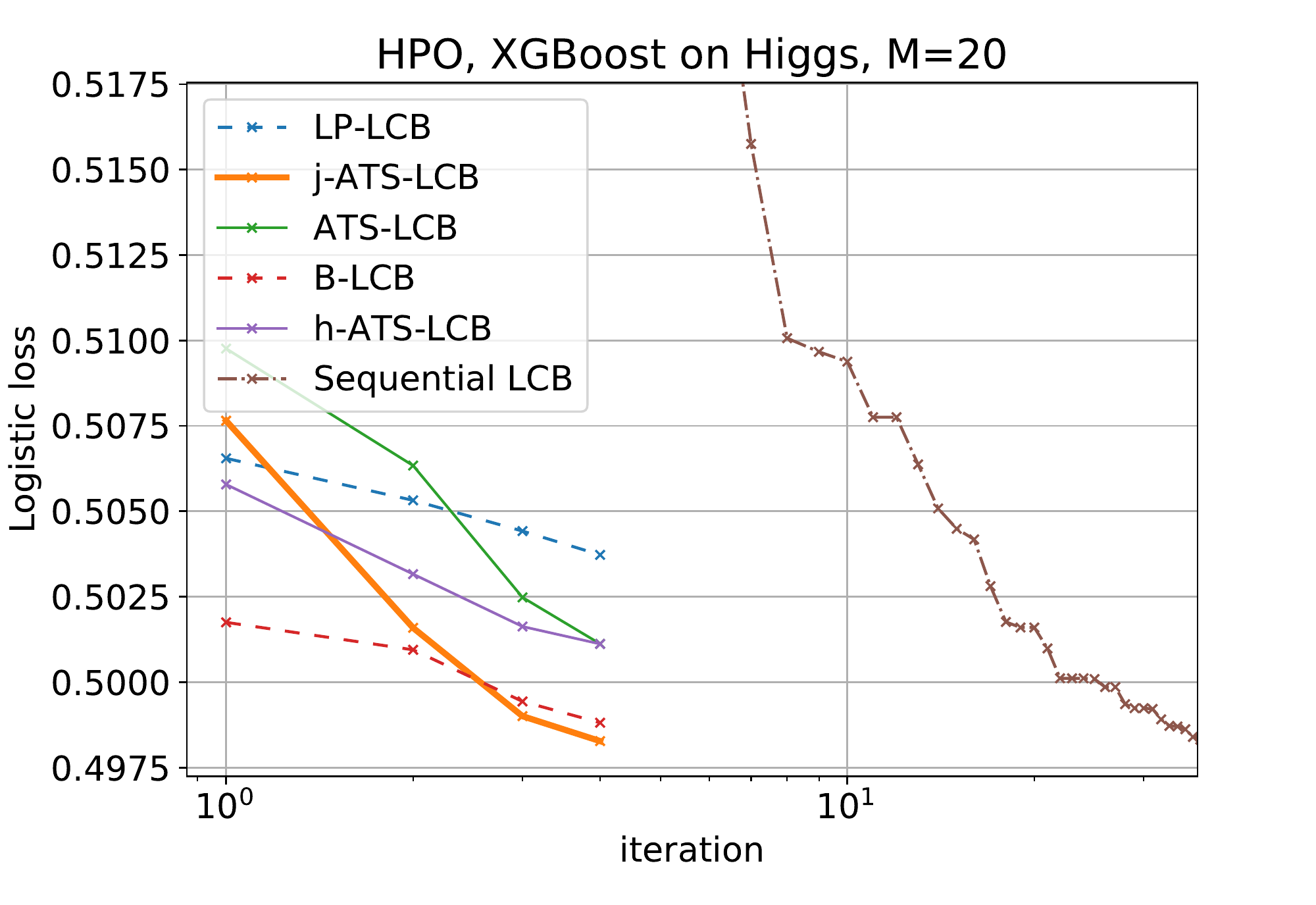} 
	\vspace{-10pt} \captionof{figure}{Hyper-parameter tuning of XGBoost on the Higgs dataset, with batch size $M=20$. }
	\label{fig:xgb-20-iters}
\end{figure}

We benchmarked against two state-of-the-art batch BO algorithms (B-LCB, LP) \citep{Gonzalez, GP-BUCB} and against Parallel Thompson Sampling \citep{kandasamy18a} on five synthetic functions from the global optimization literature, and on tuning a popular gradient boosting tree algorithm.
While ATS should be seen as orthogonal to the existing parallel methods, as it can be employed to increase their iteration efficiency, we show that the variability induced by sampling surrogate model hyper-parameters suffices to create enough parallelism to be competitive with, or to outperform, the employed parallel baselines on strongly multi-modal functions. Moreover, enhancing B-LCB and P-TS is particularly effective on objective functions exhibiting deep valleys.

\bibliographystyle{plainnat}
\bibliography{snap_hpo_bib}

\cleardoublepage
\begin{appendix}

\section{Batch Diversity in ATS}

This Appendix provides additional details on the two techniques, discussed in Section \ref{sec:j-h-ats}, for achieving batch diversity within the ATS  framework.

\subsection{$j$-ATS}

The pseudo-code for a batch iteration of the $j$-ATS variant of our algorithm is given in Algorithm \ref{alg:jitter-thompson-mcmc}.

\begin{algorithm}[h]
	\caption{Batch iteration, jittered Acquisition Thompson Sampling (j-ATS).}\label{alg:jitter-thompson-mcmc}
	\begin{algorithmic}[1]

		\Require dataset $\mathcal{D}_{t}$, sequential acquisition function $a(\boldsymbol{x}, \boldsymbol{\theta}| \mathcal{D}_t)$, batch size $M$.
		\For{$i = 1, \dots, M$ \textbf{in parallel}}
		\State \parbox[t]{\dimexpr\linewidth-\algorithmicindent} {sample $s$ new GP hyper-parameter vectors according to $\boldsymbol{\theta}_q \sim p(\boldsymbol{\theta} | \mathcal{D}_t)$. \strut}
		\State \parbox[t]{\dimexpr\linewidth-\algorithmicindent} {sample the jitter according to $j \sim p(j)$. \strut}
		\State \parbox[t]{\dimexpr\linewidth-\algorithmicindent} {$\tilde{a}_{s, j}(\boldsymbol{x} | \mathcal{D}_t) \leftarrow \frac{1}{s} \sum_{q=1}^{s} a(\boldsymbol{x}, \boldsymbol{\theta}_q, j | \mathcal{D}_t)$. \strut}
		\State \parbox[t]{\dimexpr\linewidth-\algorithmicindent} {$\boldsymbol{x}_{t+i} \leftarrow \argmax_{\mathcal{X}} \tilde{a}_{s, j}(\boldsymbol{x} | \mathcal{D}_t)$, the $i$-th batch point. \strut}
		\State \parbox[t]{\dimexpr\linewidth-\algorithmicindent} {evaluate $f(\boldsymbol{x}_{t+i})$. \strut}
		\EndFor

		\State $\mathcal{D}_{t + M} \leftarrow \mathcal{D}_{t} \cup \{(\boldsymbol{x}_k, f(\boldsymbol{x}_k))\ \forall\ t < k \leq t+M\}\ $.

	\end{algorithmic}

\end{algorithm}

\subsection{$h$-ATS}\label{app:hallucination}

As explained in Section \ref{sec:related-parallel}, approaches for parallel BO like \citep{GP-BUCB} work by augmenting the dataset with \emph{hallucinations} of the pending evaluations. Evaluation $i$ is then selected by having $a(\boldsymbol{x})$ make use of the hallucinated surrogate model $\mathcal{N}(\boldsymbol{\mu}, K\ |\ \tilde{\mathcal{D}}_{i-1})$ with $\tilde{\mathcal{D}}_{i-1} = \mathcal{D}_t\ \cup\ \{(\boldsymbol{x}_k', h_k)\ \forall\ t<k \leq i-1\}\ $, where $t$ is the time of the last completed evaluation and $h_k$ denotes a hallucination of $f(\boldsymbol{x}_k')$. As the $h_k$'s are approximations resulting from the last updated surrogate model, these approaches might be over-confident on the collected observations \citep{DPP}.

If we perform the sampling of the acquisition functions in a sequential manner, we can exploit the same technique in our randomized framework by hallucinating the data posterior, rather than the surrogate model (which remains independent of the pending evaluations). By doing that, we can make use of the knowledge of the locations of the pending evaluations in order to "update" the stochastic process to $a_s(\boldsymbol{x} | \tilde{\mathcal{D}}_{i-1})$. This means that, at iteration $i$, we sample:
\begin{equation}
\label{eq:h-as}
\tilde{a}_{s}^h(\boldsymbol{x}) = \frac{1}{s} \sum_{q=1}^{s} a(\boldsymbol{x}; \boldsymbol{\theta}_q) \text{ s.t. } \boldsymbol{\theta}_q \sim p(\boldsymbol{\theta} | \tilde{\mathcal{D}}_{i-1}).
\end{equation}
From the high-level perspective, sampling from the hallucinated data posterior means sampling models that should explain the hallucinations. In fact, the difference between $p(\boldsymbol{\theta} | \tilde{\mathcal{D}}_{t+i-1})$ and $p(\boldsymbol{\theta} | \mathcal{D}_{t})$ lies in that the former takes the hallucinations into account in its data likelihood $p(\mathcal{D}_{t}\ \cup\ \{(\boldsymbol{x}_k', h_k)\ \forall\ t<k \leq i-1\} | \boldsymbol{\theta})$. The practical effect, analogously to $j$-ATS, is to increase the diversity across the batch points. In this case, though, no acquisition function-specific jitter prior needs to be devised; the approach is therefore more readily applicable to new acquisition functions.
For what concerns hallucinations, we adapt the approximation employed by \citet{GP-BUCB} to make use of the $\boldsymbol{\theta}_q$ samples:
$h_i = \frac{1}{s} \sum_{q=1}^{s} \mu(\boldsymbol{x}_i'; \boldsymbol{\theta}_{q} | \mathcal{D}_t)$. We point out that the $\boldsymbol{\theta}_{q}$ vectors are those sampled using the hallucinated data posterior, as in \eqref{eq:h-as}, but the posterior is on the last updated model (without hallucinations). We summarize a batch iteration of this \emph{hallucinated} ATS ($h$-ATS) in Algorithm \ref{alg:hallucinated-thompson-mcmc}.

\begin{algorithm}[t]
	\caption{Batch iteration, hallucinated Acquisition Thompson Sampling (h-ATS).}\label{alg:hallucinated-thompson-mcmc}
	\begin{algorithmic}[1]

		\Require dataset $\mathcal{D}_{t}$, sequential acquisition function $a(\boldsymbol{x}, \boldsymbol{\theta}| \mathcal{D}_t)$, batch size $M$.

		\State \parbox[t]{\dimexpr\linewidth-\algorithmicindent} {$\tilde{\mathcal{D}}_{t} := \mathcal{D}_{t}$. \strut}

		\For{$i = 1, \dots, M$}
		\State \parbox[t]{\dimexpr\linewidth-\algorithmicindent} {sample $s$ new GP hyper-parameter vectors according to $\boldsymbol{\theta}_q \sim p(\boldsymbol{\theta} | \tilde{\mathcal{D}}_{t+i-1})$, the hallucinated data posterior. \strut}
		\State \parbox[t]{\dimexpr\linewidth-\algorithmicindent} {$\tilde{a}_s^h(\boldsymbol{x} | \mathcal{D}_t) \leftarrow \frac{1}{s} \sum_{q=1}^{s} a(\boldsymbol{x}, \boldsymbol{\theta}_q | \mathcal{D}_t)$. \strut}
		\State \parbox[t]{\dimexpr\linewidth-\algorithmicindent} {$\boldsymbol{x}_{t+i} \leftarrow \argmax_{\mathcal{X}} \tilde{a}_s^h(\boldsymbol{x} | \mathcal{D}_t)$, the $i$-th batch point. \strut}
		\State \parbox[t]{\dimexpr\linewidth-\algorithmicindent} {$h_{t+i} = \frac{1}{s} \sum_{q=1}^{s} \mu(\boldsymbol{x}_{t+i}, \boldsymbol{\theta}_{q} | \mathcal{D}_t)$. \strut}
		\State \parbox[t]{\dimexpr\linewidth-\algorithmicindent} {$\tilde{\mathcal{D}}_{t+i} \leftarrow \tilde{\mathcal{D}}_{t+i-1} \cup \{(\boldsymbol{x}_{t+i}, h_{t+i})\}$. \strut}
		\EndFor

		\For{$i = 1, \dots, M$ \textbf{in parallel}}
		\State \parbox[t]{\dimexpr\linewidth-\algorithmicindent} {evaluate $f(\boldsymbol{x}_{t+i})$. \strut}
		\EndFor

		\State $\mathcal{D}_{t + M} \leftarrow \mathcal{D}_{t} \cup \{(\boldsymbol{x}_k, f(\boldsymbol{x}_k))\ \forall\ t < k \leq t+M\}$.

	\end{algorithmic}

\end{algorithm}

\section{Data posterior sampling and implementation details} \label{app:details}

All ATS variants sample acquisition functions from a process defined through the hyper-parameters data posterior. As outlined in Algorithms \ref{alg:thompson-mcmc} and \ref{alg:hallucinated-thompson-mcmc}, for ATS and $h$-ATS, it suffices to sample $\boldsymbol{\theta}$, the surrogate model hyper-parameters, to obtain an acquisition function (Equation \eqref{eq:as}). $\boldsymbol{\theta}$ consists of $\boldsymbol{\theta}_k$, the GP kernel hyper-parameters, and a constant prior mean $\mu_p$ (cf. Section \ref{sec:related-sequential}). Making use of the Bayes rule, we can write:
\[ p(\boldsymbol{\theta} | \mathcal{D}_t) = \frac{1}{Z}\ p(\mathcal{D}_t | \boldsymbol{\theta})\ p(\boldsymbol{\theta}),\]
where $\frac{1}{Z}$ is a normalization constant. $p(\mathcal{D}_t | \boldsymbol{\theta})$ is the marginal likelihood of the collected data under the Gaussian process \citep{GPBook}, while $p(\boldsymbol{\theta})$ is the prior over the model hyper-parameters.
For what concerns the prior, we explored different options and eventually chose a modification of that employed by \citet{gpyopt2016}, as it was the one that performed the best in a variety of cases. Specifically:
\begin{align}
p(\boldsymbol{\theta}_{k}) &= \Gamma(\alpha=1, \beta=6)\\
p(\mu_p) &= \text{Uniform}(-3, 3).
\end{align}
The priors are independent as in \citet{spearmint}.
As we found the choice of the prior to be strongly dependent on the output range of $f$, we employ z-normalization of the outputs $\{y_i\}$ after each model update.

To sample from $p(\boldsymbol{\theta} | \mathcal{D}_t)$, we employ Affine Invariant MCMC Ensemble sampling by \citet{goodman2010}. Its computational cost is comparable to that of single-particle algorithms like Metropolis sampling \citep{goodman2010}. For BO, the main bottleneck in evaluating $p(\boldsymbol{\theta} | \mathcal{D}_t)$ resides in the matrix inversion to compute the GP marginal likelihood, cubic in the size of the samples \citep{spearmint}. Our parallel algorithm hence does not add any computational cost to a sequential Bayesian treatment of hyper-parameters: it suffices to retain more $\boldsymbol{\theta}$ samples.

\begin{table}[t!]
	\small
	\label{table}
	\centering
		\begin{tabular}{llcc}
			\toprule
			$f(\boldsymbol{x})$     & $d$  & $\mathcal{X}$ & $\min_{\mathcal{X}} f(\boldsymbol{x})$ \\
			\midrule
			Branin & $2$ & $[-5, 10] \times [0, 15]$ & $0.3979$ \\
			Cosines & $2$ & $[0, 1]^2$ & $-1.773$ \\
			Hartmann6 & $6$ & $[0, 1]^6$ & $-3.322$\\
			Eggholder & $2$ & $[-512, 512]^2$ & $-959.64$\\
			Rosenbrock4 & $4$ & $[-5, 10]^4$ & $0$\\
			\bottomrule
		\end{tabular}
	\vspace{5pt}
	\captionof{table}{Synthetic functions characteristics.}
	\label{table-benchmark-functions}
\end{table}

For $j$-ATS, the hyper-parameters also include the jitter $j$, a parameter of the acquisition function itself.
The data posterior defining the process is hence $p(\boldsymbol{\theta}, j | \mathcal{D}_t)$. As the data likelihood is independent of the parameters that do not belong to the surrogate model, we only need to define the prior of $j$. We then sample from this prior independently from the $\boldsymbol{\theta}$ sampling (see Algorithm \ref{alg:jitter-thompson-mcmc}). Let $C$ be a Bernoulli random variable with $p=0.5$. Then:
\begin{align}
\text{EI:	}&  j|C=1 \sim \text{logUniform}(-3, 0)\text{; } \qquad\\
\text{LCB:	} &j|C=1 \sim \text{Beta}(1, 12)
\end{align}
while we revert to ATS, as presented in Section \ref{sec:thompson-mcmc} (without jitter), if $C=0$. Reverting to ATS means that $j=0$ (no exploration incentive) for EI, and $j=1$ for LCB. In the latter case, we treated $j$ as a hyper-parameter to be tuned and employed the same value for B-LCB.

\section{Experiments details} \label{app:experiments}

Finally, we provide additional details for the functions we optimize in the experiments presented in Section \ref{sec:experiments}.
Table \ref{table-benchmark-functions} summarizes the characteristics of the employed synthetic functions.

Table \ref{table:hpo} reports the domains for the $5$ XGBoost hyper-parameters we tuned for the experiments in Section \ref{sec:xgb-res}; we kept the maximum number of bins equal to $64$.

\begin{table}[t!]
	\small
	\centering
	\begin{tabular}{ll}
		\toprule
		Hyper-parameter     & domain \\
		\midrule
		Number of boosting rounds &  $[16, 256]$  \\
		Step shrinkage ($\eta$) &  $[0.1, 1]$ \\
		Maximum depth of a tree  &  $[2, 8]$ \\
		Regularization coefficient ($\lambda$) &  $[10^{-2}, 10^{7}]$ \\
		Subsample ratio of columns &  $[0.1, 1]$\\
		\bottomrule
	\end{tabular} \vspace{3pt}
	\captionof{table}{Hyper-parameter domains for the tuning of XGBoost.}
	\label{table:hpo}
\end{table}

\end{appendix}

\end{document}